\pdfoutput=1

\documentclass[11pt]{article}

\usepackage[final]{acl}

\usepackage{times}
\usepackage{latexsym}

\usepackage[T1]{fontenc}
\usepackage{makecell}

\usepackage[utf8]{inputenc}

\usepackage{graphicx}
\graphicspath{ {./images/} }
\usepackage{enumitem}
\usepackage{hyperref}
\usepackage{booktabs}
\usepackage{subcaption} 
\usepackage{longtable} 
\usepackage{array}
\usepackage{float}

\title{
Team Unibuc - NLP at SemEval-2025 Task 11: 
Few-shot text-based emotion detection
}

\author{
Teodor-George Marchitan$^{1, 3}$,
Claudiu Creanga$^{2, 3}$, Liviu P. Dinu$^{1, 3}$\\ 
  $^1$ Faculty of Mathematics and Computer Science, \\
  $^2$ Interdisciplinary School of Doctoral Studies, \\
  $^3$ HLT Research Center, \\
  University of Bucharest, Romania\\
  \small
{\tt teodor.marchitan@s.unibuc.ro, ccreanga@fmi.unibuc.ro, ldinu@fmi.unibuc.ro}  \\
}

\vspace{2mm}
  
\begin{document}

\maketitle

\begin{abstract}

This paper describes the approach of the Unibuc - NLP team in tackling the SemEval 2025 Workshop, Task 11: Bridging the Gap in Text-Based Emotion Detection. We mainly focused on experiments using large language models (Gemini, Qwen, DeepSeek) with either few-shot prompting or fine-tuning. With our final system, for the multi-label emotion detection track (track A), we got an F1-macro of $0.7546$ (26/96 teams) for the English subset, $0.1727$ (35/36 teams) for the Portuguese (Mozambican) subset and $0.325$ (\textbf{1}/31 teams) for the Emakhuwa subset.

\end{abstract}

\section{Introduction}
Task 11 from the International Workshop on Semantic Evaluation \cite{muhammad-etal-2025-semeval} focuses on identifying emotions that most people would think the speaker might feel given a short piece of text. With all the advances in the AI world, automatic emotion detection plays such an important role: it can lead to more empathetic and context-aware interactions between humans and chatbots / AI assistants, improving user experience and satisfaction; it can help monitoring mental health conditions and provide timely interventions such that more people seek for help before is too late; it can help business to understand the customers sentiments regarding different products or services and can improve their strategies.

The best system developed for Task 11 track A is based on Gemini Flash \cite{geminiteam2024gemini15unlockingmultimodal} model using different techniques of few-shot prompting. 

We made our models publicly available in a \href{https://github.com/TooHappy22/emotion-detection-semeval-2025}{GitHub Repository}.

\section{Background}

The competition is multilingual and it had 3 tracks:
\begin{enumerate}[label=\Alph*.]
    \item Multi-label Emotion Detection ($28$ supported languages)
    \item Emotion Intensity ($12$ supported languages)
    \item Cross-lingual Emotion Detection ($28$ supported languages)
\end{enumerate}

We participated to track A with our system for 3 languages: Emakhuwa $0.325$ F1-macro (position \textbf{1}/31 teams), English $0.7546$ F1-macro (position 26/96 teams) and Portuguese (Mozambican) $0.1727$ F1-macro (position 35/36 teams).

\subsection{Dataset}

The BRIGHTER dataset \cite{muhammad2025brighterbridginggaphumanannotated} was collected from 4 different data sources (social media posts, personal narratives, literary texts, and news data) in $28$ different languages:

\begin{table}[!htp]
\begin{center}
\begin{tabular}{l|c|c|c}
\hline
    \thead{Language} & \thead{Train} & \thead{Dev} & \thead{Test}  \\  \hline
    \texttt{ENG} & $2,768$ & $116$ & $2,767$ \\
    \texttt{PTMZ} & $1,546$ & $257$ & $776$  \\
    \texttt{VMW} & $1,551$ & $258$ & $777$  \\
\hline
\end{tabular}
\end{center}
  \caption{Train/Dev/Test splits for languages tackled with our system for track A}
  \label{table:dataset_sizes}
\end{table}


\begin{figure}[htbp]
    \centering
    \includegraphics[width=0.9\linewidth]{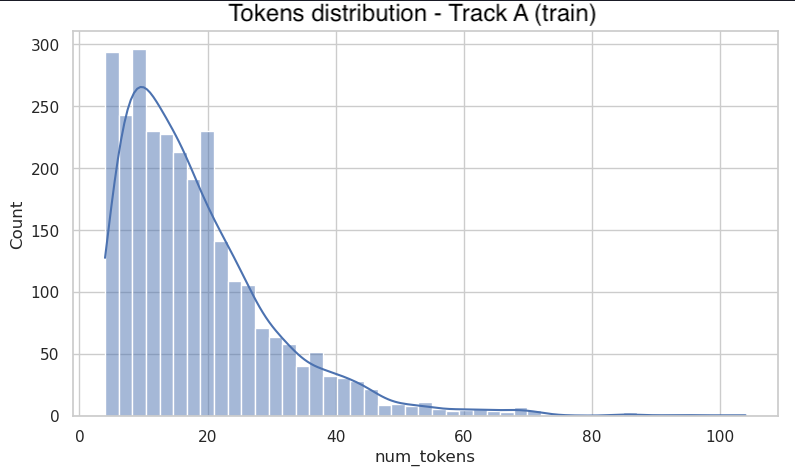}
    \caption{Track A: Distribution of token length for the training dataset in english}
    \label{fig:tokens_train}
\end{figure}

\subsection{Previous Work}

Emotion detection in text has been a subject of study for a considerable time, with early approaches relying on lexicon-based methods. These methods leverage pre-compiled lists of words associated with different emotions. For instance, the presence of words like "happy," "joyful," or "excited" might indicate a positive emotion, while words like "sad," "depressed," or "miserable" could suggest sadness. Later, traditional machine learning classifiers such as Support Vector Machines (SVMs), Naive Bayes, and Maximum Entropy models became popular for emotion detection. These models typically rely on handcrafted features extracted from text, including bag-of-words, TF-IDF, n-grams, and sentiment lexicons.  

More recently, Recurrent Neural Networks (RNNs), particularly Long Short-Term Memory networks (LSTMs) and Gated Recurrent Units (GRUs), proved effective in capturing sequential information in text, enabling models to better understand context and dependencies within sentences and documents \cite{poria-etal-2017-context}. Convolutional Neural Networks (CNNs) were also employed to extract local features and patterns indicative of emotion \cite{kim-2014-convolutional}. 

Transformer-based models, such as BERT \cite{devlin2018bert}, RoBERTa \cite{liu2019roberta}, and their variants, have achieved state-of-the-art performance in various NLP tasks, including emotion detection. These models, pre-trained on massive amounts of text data, capture rich contextual representations of words and sentences, leading to significant improvements in accuracy and robustness. Fine-tuning these pre-trained models on emotion-annotated datasets became a common practice for achieving high performance in emotion detection tasks. The SemEval challenges have consistently played a significant role in driving research in emotion detection. SemEval 2024 contained Task 10 which tackled emotion discovery and reasoning flip in conversation (EDiReF). The majority of participants used LLMs and achieved best results \cite{semeval2024}.

\section{System overview}

In this paper, we focused our research on three approaches: \textbf{fine-tuning BERT} based models  (\ref{subsec:bert}), and two techniques for LLMs: \textbf{Few-Shot Prompting} (\ref{subsec:few_shot_prompting}) and  \textbf{Fine-tuning} (\ref{subsec:fine_tuning}).

\subsection{Fine-tuning BERT based models}\label{subsec:bert}

We experimented with several prominent transformer-based models: DeBERTa v3 Large \cite{he2021deberta}, mBERT (Multilingual BERT) \cite{devlin2018bert}, and XLM-RoBERTa-large \cite{conneau2020unsupervised}. These models represent advancements in pre-trained language model architectures and have demonstrated strong performance in cross-lingual understanding tasks. We utilized these pre-trained models as feature extractors and appended a simple classification head on top. We extract sentence-level feature representations by specifically using the [CLS] token output from the transformer's final hidden layer. To prevent overfitting, we apply a dropout rate of 0.3 to these features.  These extracted features are then fed into a fully connected network, designed for classification. This network comprises two core blocks. Each block progressively reduces the feature dimensionality, starting with 512 neurons and narrowing down to 128. The final layer of the fully connected network has 6 output neurons, corresponding to the 6 emotion categories we are predicting.  We apply a higher dropout rate of 0.5 within the fully connected network, again for regularization. The model outputs raw prediction scores without any activation function.

We trained this model over 3 epochs in a two-stage approach, as recommended in \cite{marchitan-etal}. Initially, for the first 2 epochs, we froze the transformer layers, only training the weights of our fully connected classification network. This allows the classification layers to adapt to the pre-trained transformer features without disrupting the transformer's learned representations. In the final 1 epoch, we unfroze the last transformer layer and fine-tuned it along with the fully connected network. This enables the model to slightly adjust the transformer's understanding of the input text to be even more relevant for our specific emotion detection task. We used a batch size of 32 during training and employed the AdamW optimizer, known for its effectiveness in training transformers.  For the initial frozen-transformer training phase, we used a learning rate of 3e-4, which was reduced to 2e-4 during fine-tuning to prevent destabilizing the already partially trained model.  A linear learning rate scheduler with a 50-step warm-up was used to gradually increase the learning rate at the beginning of training, promoting stable convergence.  We used cross-entropy loss as our objective function.

While our BERT-based models, provided a good baseline, their performance did not reach the levels achieved by LLMs like Gemini Flash (\autoref{table:bert_models}). However, it's important to note that these BERT models offer a significant advantage in terms of computational resources. They are considerably smaller in model size and demonstrate substantially faster inference speeds compared to LLMs.

\begin{table}[H]
\centering
\begin{tabular}{l|l|c|c}
\hline
    \thead{Lang} & \thead{Model} & \thead{F1 Macro}  & \thead{F1 Micro} \\
    \hline
    \texttt{ENG} & mBERT & $0.54$ & $0.56$ \\
    \texttt{ENG} & DeBERTa & $0.70$ & $0.70$ \\
    \texttt{ENG} & XLM-RoBERTa & $0.70$ & $0.71$ \\
\hline
\end{tabular}
  \caption{Results of BERT based models for validation set.}
  \label{table:bert_models}
\end{table}

\subsection{Few-Shot Prompting}\label{subsec:few_shot_prompting}

We found that Few-Shot prompting beats Zero-Shot by around 7\% in performance. Prompts can be inspected in \autoref{sec:prompts}. For example Qwen2.5-14B Zero-Shot CoT obtained an F1 Macro of 0.63, while with Few-Shot 0.70. This was observed across all models. Another insight was that increasing the number of examples we gave to the model will result in an increase in performance. We tried giving 6 examples, one that included each emotion and progresively giving it more examples: 100, 300 and 600. Best results were when we gave the most examples (600). A third insight is that we found out that complex prompting techniques like CoT and ToT actually make the models perform worse.  We believe this is because the task is a perceived emotion task, where the annotations are not necesarily the ground truth, but what the labelers considered to be true. Following the thinking process of the models, in the examples where they disagreed with the ground truth, we were actually conviced that the models were right in most of the cases. 

Our experiments (\autoref{table:few_shot}) revealed a consistent performance advantage for few-shot prompting over zero-shot prompting, with an approximate improvement of 7\% in F1-macro. For example, the Qwen2.5-14B model \cite{qwen2025qwen25technicalreport} achieved an F1-macro score of 0.63 using zero-shot prompting, while few-shot prompting boosted performance to 0.70. This trend was observed across all models evaluated.  Furthermore, we investigated the impact of increasing the number of examples provided within the few-shot prompts.  By progressively increasing the example count (from an initial set of 6 examples, one for each emotion, to 100, 300, and finally 600 examples), we observed a positive correlation between the number of examples and performance. The highest F1-macro scores were consistently obtained when utilizing the largest example set of 600.

Counterintuitively, we found that employing more complex prompting techniques such as Chain-of-Thought (CoT) \cite{wei2022} and Tree-of-Thought (ToT) \cite{long2023tree} did not yield the expected performance gains. In fact, these methods tended to slightly degrade performance in our specific task. We hypothesize that this unexpected outcome stems from the nature of the "perceived emotion" task itself. In this task, annotations do not necessarily represent an objective "ground truth" emotion, but rather reflect human annotators' subjective interpretations of the speaker's likely feelings. Consequently, while CoT and ToT are designed to encourage models to mimic human-like reasoning processes, in this context, forcing the model to explicitly articulate a detailed thought process may not align with the inherently subjective and potentially less consciously reasoned nature of perceived emotion annotation.  Indeed, in instances where model predictions diverged from the assigned labels, our qualitative analysis suggested that the models' reasoning, often aligned with alternative, yet plausible, emotional interpretations of the text.

\begin{table}[H]
\centering
\begin{tabular}{l|l|c|c}
\hline
    \thead{Model} & \thead{Type} & \thead{Examples}  & \thead{F1 Macro} \\
    \hline
    \texttt{Qwen2.5} & Zero-Shot & $0$ & $0.63$ \\
    \hline
    \texttt{Qwen2.5} & Z-S CoT & $0$ & $0.63$ \\
    \hline
    \texttt{Gemini} & F-S ToT & $6$ & $0.63$ \\
    \hline
    \texttt{Gemini} & F-S ToT & $20$ & $0.64$ \\
    \hline
    \texttt{Gemini} & Few-Shot & $6$ & $0.69$ \\
    \hline
    \texttt{Qwen2.5} & Few-Shot & $6$ & $0.70$ \\
    \hline
    \texttt{Gemini} & Few-Shot & $500$ & $0.76$ \\
    \hline
    \texttt{Gemini} & Few-Shot & $600$ & $0.77$ \\
\hline
\end{tabular}
  \caption{Results of LLM models for validation set. The exemples were given from the training set. We used Qwen2.5-14B and Gemini 2.0 Flash Exp. Z-S means Zero-Shot, F-S means Few-Shot.}
  \label{table:few_shot}
\end{table}

\subsection{Fine-tuning}\label{subsec:fine_tuning}

We experimented fine-tuning with different large language models (DeepSeek \cite{deepseekai2025deepseekr1incentivizingreasoningcapability}, Mistral 7B \cite{jiang2023mistral7b} and Qwen2.5 \cite{qwen2025qwen25technicalreport}) as the backbone of our system architecture, on top of which we added a multi-label classification layer in order to classify the emotions present in the given text. We set the maximum number of tokens to $256$ and we truncated the longer text by keeping the first part of the text, as suggested in \cite{marchitan-etal}. We then fine-tuned the quanitized model using Low-Rank Adaptation (LoRA) \cite{hu2022lora} with following hyperparameters: $r=4$ ($r=2$ for Mistral 7B), $lora\_alpha=8$ and $lora\_dropout=0.05$ ($lora\_dropout=0.1$ for Qwen2.5-1.5B). The fine-tuning was done using the AdamW optimizer \cite{loshchilov2019decoupledweightdecayregularization} with a weight decay of $0.01$, weight ratio of $1\%$ for Mistral 7B and Qwen2.5-1.5B, but $2\%$ for other 2 models. We set the learning rate to $2e-5$ when fine-tuning DeepSeek and Qwen2.5-1.5B, $5e-5$ for Mistral 7B and $2e-4$ for Qwen2.5-0.5B. Cross entropy loss was used during training to measure the performance. We used different batch sizes ($4$ for Mistral 7B; $8$ for DeepSeek and Qwen2.5-0.5B; $32$ for Qwen2.5-1.5B) based on the total number of trainable parameters in order to be able to fit on the available hardware.  We set the maximum number of epochs to $15$, but the actual number of training epochs varies as we implemented the early stopping strategy and the final model is the one with the smallest loss on the validation set.

\begin{table}[htp]
\begin{tabular}{l|c|c|c}
\hline
    \thead{LLM Backbone} & \thead{Epochs} & \thead{Train Loss} & \thead{Dev Loss} \\
    \hline
    \begin{tabular}{@{}c@{}}DeepSeek R1 \\ Distill \\ Llama 8B\end{tabular} & $5$ & $0.2495$ & $0.3181$ \\
    \hline
    Mistral 7B & $3$  & $0.2515$ & $0.3455$ \\
    \hline
    Qwen2.5-0.5B & $3$  & $0.3941$ & $0.3614$ \\
    \hline
    Qwen2.5-1.5B & $8$  & $0.3714$ & $0.3571$ \\
\hline
\end{tabular}
  \caption{Train and validation losses for each model alongside the number of epochs trained.}
  \label{table:train_dev_losses}
\end{table}

\begin{table*}[htbp]
\centering
\begin{tabular}{l|c|c|c|c}
    \hline
    \thead{LLM Backbone} & \thead{Validation F1 Micro} & \thead{Validation F1 Macro} & \thead{Test F1 Micro} & \thead{Test F1 Macro} \\
    \hline
    \begin{tabular}{@{}c@{}}\texttt{DeepSeek R1} \\\texttt{Distill} \\\texttt{Llama 8B}\end{tabular} & $0.7723$ & $0.7602$ & $0.7744$ & $0.7441$ \\
    \hline
    \texttt{Mistral 7B} & $0.7696$  & $0.7305$ & $0.7699$  & $0.7268$ \\
    \hline
    \texttt{Qwen2.5-0.5B} & $0.7496$  & $0.6975$ & $0.7224$  & $0.6840$ \\
    \hline
    \texttt{Qwen2.5-1.5B} & $0.7340$  & $0.6906$ & $0.7319$  & $0.6826$ \\
\hline
\end{tabular}

\caption{F1 Micro and F1 Macro results on both train and validation datasets.}
\label{table:dev_test_scores}
\end{table*}




\section{Results}

We participated in Track A and tested our proposed system on 3 languages: English, Portuguese (Mozambican) and Emakhuwa. We managed to be over the baseline on 2 out of the 3 languages and secured the first position for one of them, as shown in \autoref{table:team_results}. This reflects our model’s ability to adapt with few-shot in-context learning especially on languages with limited availability of NLP resources (such as Emakhuwa).

\begin{table}[H]
\centering
\begin{tabular}{l|c|c}
\hline
    \thead{Language Code} & \thead{F1 Macro}  & \thead{Place} \\
    \hline
    \texttt{VMW} & $0.3250$  & \textbf{1} / $31$ \\
    \texttt{ENG} & $0.7546$ & $26$ / $96$ \\
    \texttt{PTMZ} & $0.1727$ & $35$ / $36$ \\
\hline
\end{tabular}
  \caption{Team Unibuc - NLP results on Track A.}
  \label{table:team_results}
\end{table}

\subsection{Error Analysis}

\begin{figure}[htbp]
    \centering
    \includegraphics[width=1\linewidth]{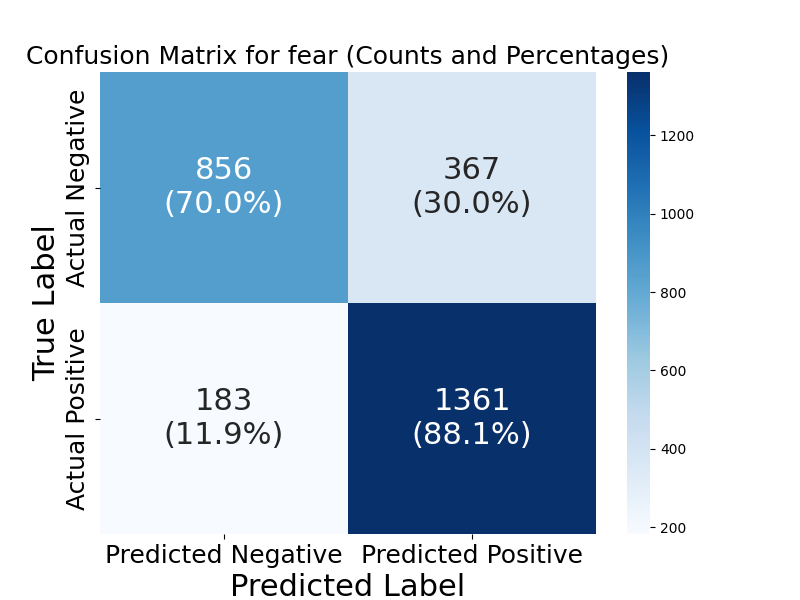} 
    \caption{Confusion Matrix for Fear (Counts and Percentages)}
    \label{fig:cm_fear}
\end{figure}

\begin{figure}[htbp]
    \centering
    \includegraphics[width=1\linewidth]{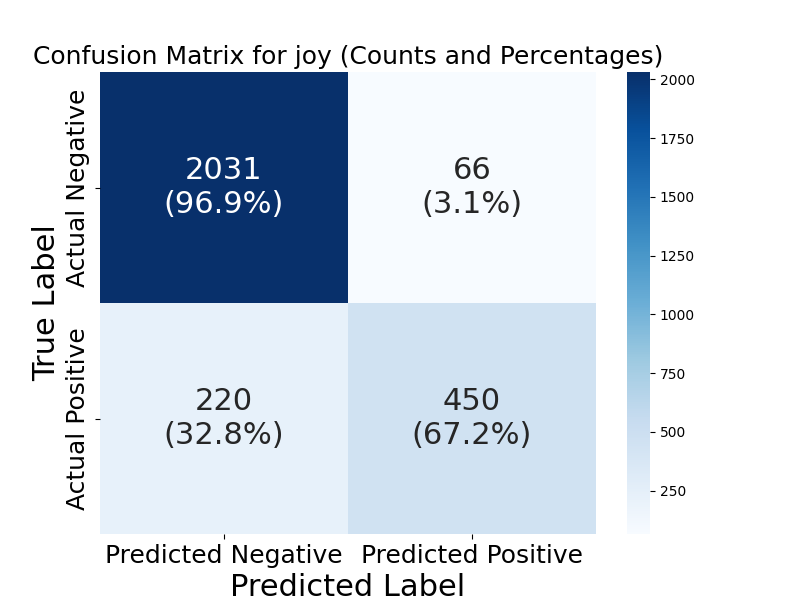} 
    \caption{Confusion Matrix for Joy (Counts and Percentages)}
    \label{fig:cm_joy}
\end{figure}

Examining the confusion matrices (\autoref{fig:cm_fear}, \autoref{fig:cm_joy},   \autoref{fig:cm_anger}, \autoref{fig:cm_sadness}, \autoref{fig:cm_surprise}), we observe distinct performance profiles across emotions. For fear, while the model achieves its highest True Positive rate (TP) at 88\%, indicating strong detection of actual fear, it simultaneously exhibits its lowest True Negative rate (TN) at 70\%. This combination suggests a tendency to over-predict fear, potentially misclassifying other emotions as fear. Conversely, for joy, the model shows the weakest performance in detecting the emotion when it is present, achieving the lowest TP of 67\% and consequently, the highest False Negative rate (FN) of 32\%, indicating a higher likelihood of missing instances of joy.  Furthermore, the highest False Positive rate (FP) of 30\% is also associated with fear, reinforcing the observation of potential over-prediction for this emotion. It seems like with both approaches we have experimented with LLMs (few-shot prompting and fine-tuning) the models tend to over-predict fear while under-predicting joy (\autoref{fig:f1_score_per_emotion}).

\section{Conclusions and Future Work}

Our investigation explored three primary approaches: fine-tuning BERT-based models and leveraging LLMs through both few-shot prompting and fine-tuning techniques. While fine-tuned BERT models provided a solid performance baseline, they were ultimately surpassed by methods employing LLMs, particularly Gemini Flash with few-shot prompting, which formed the basis of our best-performing system. Interestingly, complex prompting strategies like Chain-of-Thought and Tree-of-Thought did not yield improvements, but degradations, possibly due to the subjective nature of the perceived emotion task.

Building upon the insights gained from this study, several avenues for future research emerge.  Firstly, further investigation is warranted to understand the performance disparity observed across languages, particularly the lower F1-macro score for the Portuguese subset.  Detailed dataset analysis and error analysis could reveal language-specific nuances or data biases impacting model performance. Secondly, exploring the new "thinking models" like "Gemini 2.0 Thinking" and "Thinking Claude" should be tested, as they show promises in other similar tasks.

\section*{Acknowledgements}

Research partially supported by the Ministry of Research, Innovation and Digitization, CNCS/CCCDI UEFISCDI, SiRoLa project, PN-IV-P1-PCE-2023-1701, and InstRead project PN-IV-P2-2.1-TE-2023-2007, within PNCDI IV, Romania.

\bibliography{main}

\appendix

\section{Example of prompts}\label{sec:prompts}

\subsection{Zero shot}

Analyze the following sentence and identify all emotions that are present. Select from this list: "Anger", "Fear", "Joy", "Sadness", "Surprise". If multiple emotions are present, list them separated by commas. If no emotions from the list are present, respond with "None".

Sentence: [Sentence]

Emotions:

\subsection{Zero shot with CoT}

Analyze the following sentence and identify all emotions that are present. Select from this list: "Anger", "Fear", "Joy", "Sadness", "Surprise". If multiple emotions are present, list them separated by commas. If no emotions from the list are present, respond with "None". Let's break down the emotional content step by step.

Sentence: [Sentence]

Reasoning: Consider the specific words used, the context of the sentence, and any implied feelings.

Emotions:

\subsection{Few shot with CoT}

Analyze the following sentence and identify all emotions that are present. 

Examples:
1.  Sentence: "But not very happy."
    Emotions: Joy, Sadness
2.  Sentence: "They were dancing to Bolero"
    Emotions: Joy, Sadness
3.  Sentence: Yes, the Oklahoma city bombing."
    Emotions: Anger,Fear,Sadness,Surprise
4. Sentence: "5 year old me was scarred for life."
    Emotions: Fear, Sadness
5. Sentence: "How stupid of him."
    Emotions: Anger
6. Sentence: "I turned around so I could see my back."
    Emotions: Surprise
    
Given the following sentence, select from this list: "Anger", "Fear", "Joy", "Sadness", "Surprise". If multiple emotions are present, list them separated by commas. Let's break down the emotional content step by step.

Sentence:  [Sentence]

Reasoning: Consider the specific words used, the context of the sentence, and any implied feelings. Output only the emotions that are present. No other words.

Emotions:

\subsection{Few Shot with ToT}

Analyze the following sentence and identify all emotions that are present. Given the following sentence, select from this list: "Anger", "Fear", "Joy", "Sadness", "Surprise".

Examples: [600 examples]

Given the following sentence, select from this list: "Anger", "Fear", "Joy", "Sadness", "Surprise". If multiple emotions are present, list them separated by commas. Let's break down the emotional content step by step using a Tree of Thoughts approach.

Sentence: [Sentence]

Reasoning:

Thought 1: Initial Impression: What is the first emotion that comes to mind upon reading the sentence? Briefly explain why.

Thought 2: Word-Level Analysis: Are there any specific words or phrases that strongly suggest an emotion? If so, which words and which emotions?

Thought 3: Contextual Considerations: Does the context of the sentence provide any additional clues about the emotional state? Consider the situation being described.

Thought 4: Alternative Interpretations: Are there any other possible interpretations of the sentence that might suggest different emotions? Explore these possibilities.

Thought 5: Synthesis: Based on the previous thoughts, which emotions are most likely present in the sentence? Justify your final selection.

Final Emotions: Output only the emotions that are present, separated by commas. No other words.

Emotions:

\section{Error analysis}

\begin{figure}[H] 
    \centering
    \includegraphics[width=1\linewidth]{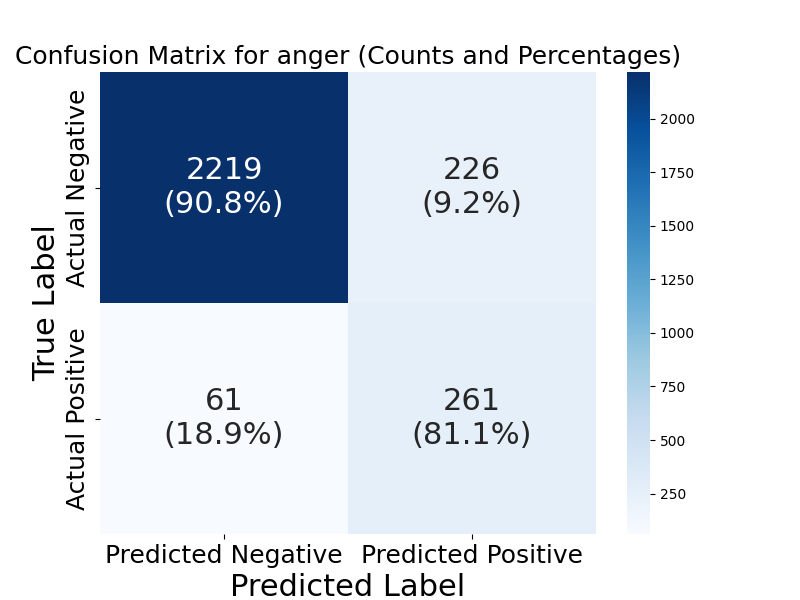} 
    \caption{Confusion Matrix for Anger (Counts and Percentages)}
    \label{fig:cm_anger} 
\end{figure}

\begin{figure}[H]
    \centering
    \includegraphics[width=1\linewidth]{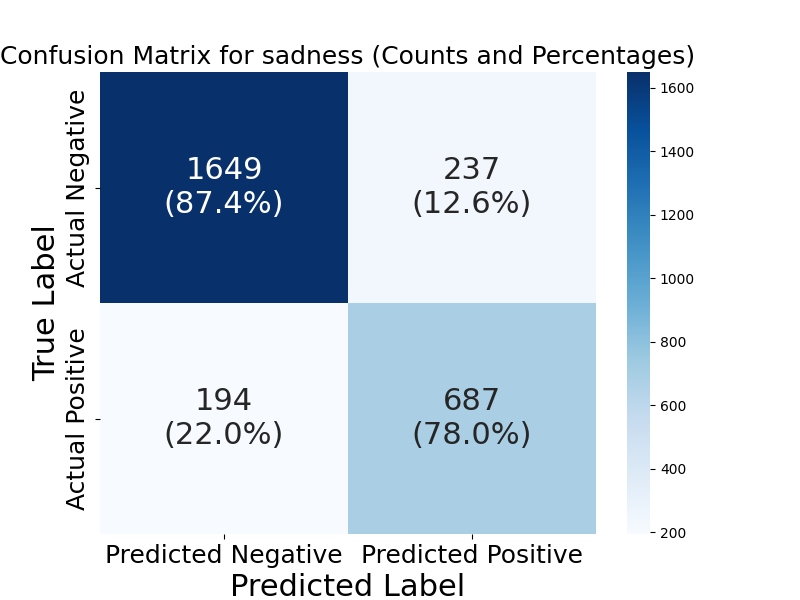} 
    \caption{Confusion Matrix for Sadness (Counts and Percentages)}
    \label{fig:cm_sadness}
\end{figure}

\begin{figure}[H]
    \centering
    \includegraphics[width=1\linewidth]{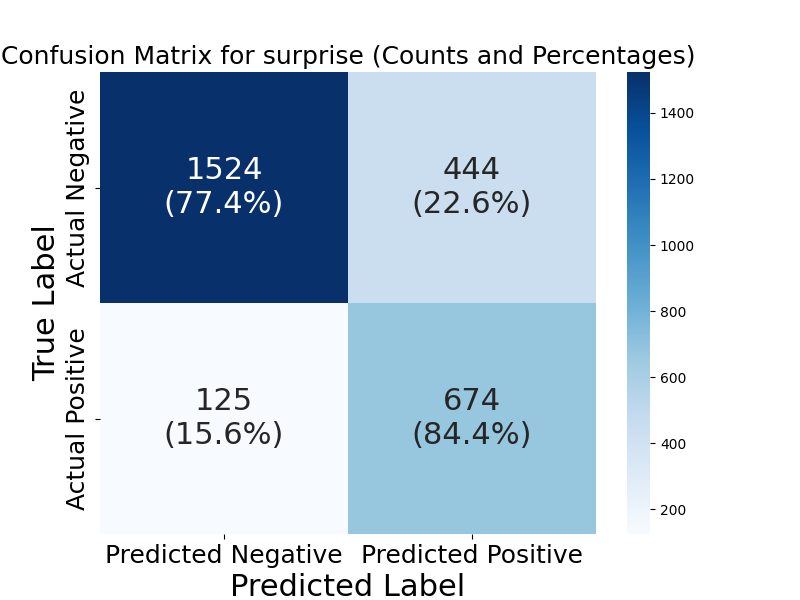} 
    \caption{Confusion Matrix for Surprise (Counts and Percentages)}
    \label{fig:cm_surprise}
\end{figure}

\begin{figure}[H]
    \centering
    \includegraphics[width=1\linewidth]{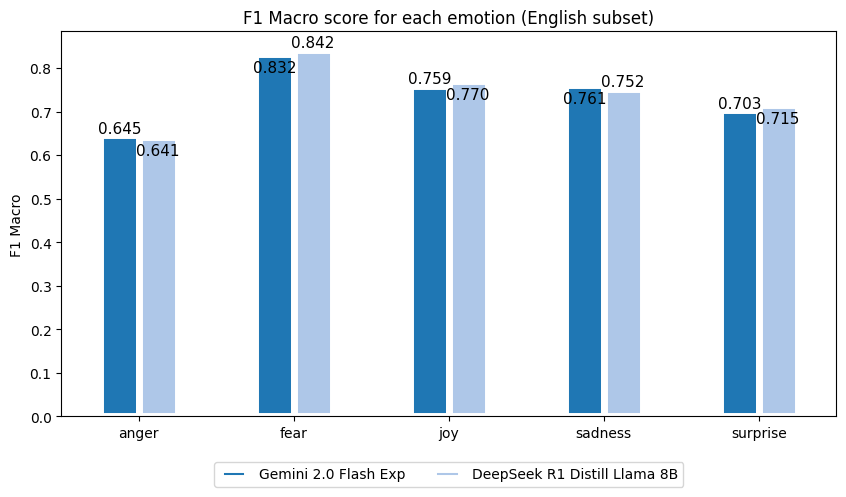} 
    \caption{Comparison of F1 macro score on each emotion between best model using few-shot prompting (Gemini 2.0 Flash) and fine-tuning (DeepSeek R1 Distill Llama 8B).}
    \label{fig:f1_score_per_emotion}
\end{figure}

\bigskip

\end{document}